# Study on the effectiveness of AutoML in detecting cardiovascular disease

T.V. Afanasieva [0000-0003-3779-7992], A.P. Kuzlyakin, A.V. Komolov

The Plekhanov Russian University of Economics, Stremyanny lane, 36, Moscow, Russia.
tv.afanasjeva@gmail.com

The corresponding author

T.V. Afanasieva, The Plekhanov Russian University of Economics, Stremyanny lane, **36**, Moscow, Russia

tv.afanasjeva@gmail.com

**Abstract.** Cardiovascular diseases are widespread among patients with chronic noncommunicable diseases and are one of the leading causes of death, including in the working age. The article presents the relevance of the development and application of patient-oriented systems, in which machine learning (ML) is a promising technology that allows predicting cardiovascular diseases. Automated machine learning (AutoML) makes it possible to simplify and speed up the process of developing AI/ML applications, which is key in the development of patient-oriented systems by application users, in particular medical specialists. The authors propose a framework for the application of automatic machine learning and three scenarios that allowed for data combining five data sets of cardiovascular disease indicators from the UCI Machine Learning Repository to investigate the effectiveness in detecting this class of diseases. The study investigated one AutoML model that used and optimized the hyperparameters of thirteen basic ML models (KNeighborsUnif, KNeighborsDist, LightGBMXT, LightGBM, RandomForestGini, RandomForestEntr, CatBoost, ExtraTreesGini, ExtraTreesEntr, NeuralNetFastA, XGBoost, NeuralNetTorch, LightGBMLarge) and included the most accurate models in the weighted ensemble. The results of the study showed that the structure of the AutoML model for detecting cardiovascular diseases depends not only on the efficiency and accuracy of the basic models used, but also on the scenarios for preprocessing the initial data, in particular, on the technique of data normalization. The comparative analysis showed that the accuracy of the AutoML model in detecting cardiovascular disease varied in the range from 87.41% to 92.3%, and the maximum accuracy was obtained when normalizing the source data into binary values, and the minimum was obtained when using the built-in AutoML technique.

## Introduction

Currently, many countries are dealing with an overloaded healthcare system and a shortage of qualified doctors, so patient-oriented systems (POS), which actively use machine learning models, show great promise. Patient-oriented systems are systems whose purpose is to empower people to improve their health with the help of digital technologies used to prevent diseases at the individual level and at home. In this aspect, the POS is an innovative tool for digital disease prevention, which has a number of advantages: improving the quality and accessibility of preventive measures, completeness of patient data, reducing the pressure on medical staff, reducing the time and material costs of preventive and curative measures. The main advantages for the population when using POS are: Engaging patients in the care of their health, improving the quality and active life expectancy, Ensuring monitoring and control of their health.

The POS includes health recommendation systems, virtual assistants, chatbots, and symptom monitors that provide self-monitoring for patients and actively use machine learning (ML) models to assess health status, early diagnosis of diseases, and forecast the probability of serious events requiring hospitalization. [1, 2, 3]

However, the development of ML models is mainly implemented by data scientists and programmers, it requires a lot of time to select ML models and their parameters. This limits their development and application in the real practice of the therapeutic and diagnostic process. Some problems in the field of ML application in healthcare are covered in the article [4]. Particularly, it is noted that there is a significant gap in the application of ML, because there are relatively few solutions in the real world. One reason in this area is that developers offer machine learning models adapted to open data, and not to solve a real clinical problem within the framework of a therapeutic and diagnostic process. The article presents the concept according to which the successful integration of ML into the provision of medical services requires considering ML as a possibility of a wider set of technologies and workflows, rather than the final product itself.

To solve the above-mentioned problems, models are being developed that allow automating the process of developing ML models as a POS component based on the Automated machine learning (AutoML) approach. The value of AutoML models lies in its ability to democratize machine learning and make it accessible to non-specialists, thereby reducing the time and resources needed to develop accurate models.

The purpose of this article is to analyze AutoML models for the detection of cardiovascular diseases and compare the effectiveness of AutoML models with ML models by the accuracy criterion.

The following research questions were formulated by the authors:

**RQ1.** Will the use of an AutoML model that includes optimized baseline ML models lead to an increase in the accuracy of detecting cardiovascular diseases compared to basic ML models?
**RQ2.** Can AutoML's built-in preprocessing algorithm compete in efficiency with data preprocessing algorithms created by a data scientist?

**Related work**

The article [5] describes the basics, the research goals of ML and its relevance in healthcare, as well as the areas of medicine where ML methods are already used (for analyzing patient data and for early diagnosis of diseases, identifying signs of an epidemic or pandemic, drug development, radiology, etc.).

A systematic review of automated diagnostics for predicting cardiovascular diseases(CVD) based on clinical signs, images and ECG is given in the article [6] and some future research directions in the field of automatic detection of heart diseases based on machine learning and several data modalities are presented. The authors considered some limitations of ML-based methods: training on a large amount of data is a complex and time-consuming task, machine learning models may suffer from the problem of data retraining, deep learning technology requires a huge amount of data to train the model, which is expensive and difficult work. Time complexity is another problem of automatic detection of heart diseases based on machine learning approaches.

The application of the multilayer perceptron (MLP) model for predicting the condition and diagnosis of patients with coronary heart disease is considered in the study [7]. The authors used normalized data from the UCI machine learning repository. Several experiments were conducted to determine the optimal network parameters, as a result, the optimized MLP system model achieved an accuracy of 92.2%, which proves its usefulness in the process of diagnosing coronary heart disease.

The successful use of machine learning requires a lot of effort from human experts, given that no algorithm can provide good performance and the necessary accuracy for all possible data.

One of the solutions in this direction is the use of AutoML models, which focus on automating the process of model selection, optimal tuning of hyperparameters and the development of features [8]. One of the key areas of attention was the automated search for neural architecture (NAS), which includes the search for the optimal architecture for a neural network [9]. It has been proven that the NAS has high efficiency, while some models provide the most up-to-date performance on various control datasets. Another area that AutoML focused on was the selection and configuration of the ML model. This includes selecting the appropriate ML model and configuring its hyperparameters to optimize performance and accuracy. Several automated methods for selecting and configuring models have been proposed, including Bayesian optimization, evolutionary algorithms, and reinforcement learning [10].

Realizing the limitations of using one ML model to detect heart disease based on the results of self-monitoring using sensors, the authors of the study [11] implemented several (Naïve Bayes, Support Vector Machine, Simple Logistic Regression, Random Forest & Artificial Neural Network (ANN)) and proposed the architecture of a web application for a patient-oriented system, in which the best ML model is automatically selected according to the accuracy criterion. The initial data for the detection of CVDs included age, blood pressure, heartbeat, gender, ECG results, blood sugar level of the patient. The model shows the probability of occurrence of CVDs. As part of the experiment, a model was defined on a database that includes Cleveland Clinic Foundation datasets that shows the probability of occurrence of CVDs with the best accuracy of 97.53%, namely, the Support Vector Machine. Using this app, the patient can enter the current heart disease parameters from anywhere on the app's interface and view the risk level of heart disease.

The authors of the article [12] compared machine learning models such as random forest, decision tree, multilayer perceptron and XGBoost for the diagnosis and prognosis of CVDs based on classification. In order to optimize the result, GridSearchCV was used to adjust the hyperparameters of the application model. The models were trained on a real dataset of 70,000 instances from Kaggle, divided in an 80:20 ratio, and achieved the following accuracy: decision tree: 86.37% (with cross-validation) and 86.53% (without cross-validation), XGBoost: 86.87% (with cross-validation) and 87.02% (without cross-validation), random forest: 87.05% (with cross-validation) and 86.92% (without cross-validation), multilayer perceptron: 87.28% (with cross-validation) and 86.94% (without cross-validation). The conclusion drawn from this underlying study is that the cross-validated multilayer perceptron outperformed all other algorithms in terms of accuracy of 87.28%.

The article [13] provides an overview of publications in the field of (AutoML) for healthcare. The main limitation of AutoML at the moment is that it is not sufficiently efficient to work with big data, that is, outside of historical datasets of small and medium size. Based on the analysis of 101 articles, the potential opportunities and obstacles for the use of AutoML in healthcare are identified, it is shown that these automated methods can match or improve human performance in certain machine learning tasks.

For example, the authors in the study [14] used AutoML to develop a deep learning model to predict the likelihood of readmission of patients with heart failure. The study [15.] showed that the use of a simple single-layer perceptron with hyperparameter tuning can implement a binary classification of patients into those who have CVDs and those who do not have them with good accuracy. In this study, hyperparameters such as hidden layer width, learning rate, and activation function by using random and grid search method are optimized. The developed simple model of the ANN on the Cleveland dataset obtained from the machine learning repository at the University of California, Irvine (UCI) achieved an accuracy of 93.44%, in addition, the time spent on training the model was reduced.

The AutoML application for identifying the risk of developing CVDs using data on 423,604 participants without CVDs at the initial stage in the British biobank is available in the article [16]. The authors compare the results of the AutoML framework (AutoPrognosis) with the traditional Framingham-score result of assessing fatal events over 10 years and with the COX PH model by criterion using the area under the receiver of the operational characteristic curve (ACC-ROC). When using the AutoML framework algorithm (auto forecast) showed the best AUC-ROC indicators: 0.774, 95% CI: 0.768-0.780.

**Proposed framework**

The proposed framework for achieving the goal set in the article, namely, the development and study of the AutoML model and its comparison with ML models according to the effectiveness criteria for the binary classification of patients into two classes in relation to the presence of CVD, includes the following stages:

1. Data preprocessing. Downloading, analyzing, cleaning and transforming data obtained as a result of combining several data sets containing information about people who provided their data on the presence of CVD.
2. Design and train an AutoML model on a training sample.
    a. training of the basic ML models included in AutoML with the adjustment of their hyperparameters,
    b. selection of ML models that have shown high accuracy on the training sample to be included in the AutoML model,
    c. training an AutoML model and calculating its accuracy on a training sample.
3. Application and evaluation of trained AutoML and ML models for the detecting of CVDs on a test sample, calculation of accuracy indicators and comparison of accuracy of AutoML and ML models. This article uses the following main assessments of the effectiveness of the detection of CVDs:

Accuracy shows the proportion of correct classifications:

$$Acc=(TP + TN)/(TP + TN + FP + FN).$$

Recall - (completeness) shows the ratio of correctly classified objects of a class to the total number of elements of this class:

$$Recall = TP/(TP + FN).$$

Precision - shows the proportion of correctly classified objects among all objects that are classified to this class:

$$Precision = TP/(TP + FP).$$

$$F\text{-measure} = 2 * Precision * Recall / (Recall + Precision).$$

In addition to the considered metrics of classification efficiency, the area under the receiver operating characteristic curve (AUC-ROC) was also used. Below we will consider the first two stages of the proposed framework in more detail, and the third stage will be consecrated in the next section.

**Stage 1. Data pre-processing**
To investigate the effectiveness and accuracy of the AutoML model, we used five datasets from the UCI Machine Learning Repository (https://archive.ics.uci.edu/ml/machine-learning-databases/heart-disease/), which were combined for 11 common indicators of CVD and formed new database of 1190 observations available today for research purposes: Clevelan (303 observations), Hungarian (294 observations), Switzerland (123 observations), Long Beach VA (200 observations), Stalog (Heart) Data Set (270 observations). Below is a list of CVD indicators, with the last sign being the target that needs to be predicted, it contains a disease class label (1: have heart disease, 0: no heart disease)
1. **Age**: Patient's age [years]
2. **Sex**: [M, F]
3. **ChestPainType**: type of chest pain [TA: tyPOSal angina, ATA: atyPOSal angina, NAP: non-anginal pain, ASY: asymptomatic]
4. **RestingBP:** resting blood pressure [mmHg]
5. **Cholesterol**: serum cholesterol level [mm/dL]
6. **FastingBS**: fasting blood sugar [1: if FastingBS > 120 mg/dL, 0: otherwise]
7. **RestingECG**: resting electrocardiogram results [Normal: normal, ST: ST-T anomaly (T wave inversion and/or ST elevation or depression > 0.05 mV), LVH: probable or definite left ventricular hypertrophy according to Estes criteria]
8. **MaxHR:** Maximum Heart Rate Achieved [Numerical value between 60 and 202]
9. **ExerciseAngina**: exercise-induced angina [Y: Yes, N: No]
10. **Oldpeak**: old peak = ST [Numerical value measured in depression]
11. **ST_Slope**: [Up, Flat, Down]
12. **HeartDisease**: Output Class [1: Heart Disease, 0: Normal]

Analyzing, cleaning, and transforming the data at the preprocessing stage will be performed using three scenarios to answer the RQ2 research question. As expected, different scenarios in the preprocessing phase of the source data can lead to the creation of different AutoML model structures that perform classification with different accuracy.
   The first S1 scenario will use the built-in capabilities of the AutoML model, which include automatic execution of time-consuming manual steps: handling missing data, manual feature transformations, data splitting [17], as well as software-implemented removal of repetitive observations and outliers in the data, if any.
   The second S2 scenario includes software-implemented data preprocessing that implements the following features:
   a. Statistical analysis of aggregated data and visualization of distributions for each CVD indicator for data quality validation

b. Data cleansing based on identifying and removing data duplications, contextual anomalies, and outliers associated with errors in data values, such as **RestingBP=0**
c. Convert cleansed data (categorical, continuous, character binary) to binary values 0 and 1. For these purposes, the values of the continuous data of each indicator will be divided into 5 intervals and each interval will be represented by a categorical value. This allows the One-Hot Encoding method to convert categorical indicators to values 0 or 1, adding additional columns to the original date set.
d. Dividing the dataset into two, for training and testing models in an 80:20 ratio.

The difference between the third S3 scenario and the second scenario is the way the cleansed data is transformed. It will apply the Z-score of continuous values to the range of real numbers [0, 1]:

$$Z(x)=(x-mean(x))/ stdev(x),$$

here mean(x) denotes the mean value, and stdev(x) is the standard deviation of the continuous exponent x.

The statistical analysis of the data, the indicators of which were included in the dataset used, revealed that the average age was 53 years (the minimum age was 28 years and the maximum age was 77 years). The majority of patients were men, the number of observations labeled "has CVD" is less than the number of observations labeled "does not have CVD", visual analysis of the data distribution confirmed that the number of patients with CVD increases with age.

After eliminating duplicates, the final set of data was 918 observations. Visual analysis of the data showed that availability outliers in Cholesterol (19% of the total number of observations) and RestingBP (4% of the total number of observations). As a result of deletion outliers Dataset for training AutoML included 714 cleaned observations, which were separated in a ratio of 80% for training and 20% for testing.

**Stage 2. Designing and training an AutoML model on a training sample**

In this article, we made a decision for AutoML to use a WeightedEnsemble_L2 model, which will be built using 13 basic first-level ML models: KNeighborsUnif, KNeighborsDist, LightGBMXT, LightGBM, RandomForestGini, RandomForestEntr, CatBoost, ExtraTreesGini, ExtraTreesEntr, NeuralNetFastA, XGBoost, NeuralNetTorch, LightGBMLarge.

A short description of each model reviewed is provided below.

1. KNeighborsUnif and KNeighborsDist:
    1.1. KNeighborsUnif: This is a k-nearest neighbors model that uses uniform distance to determine neighbors. It assigns the new data instance to the class represented by most of the neighbors in the neighborhood.
    1.2. KNeighborsDist: This is also a k-nearest neighbor model, but uses distance over a metric (e.g., Euclidean distance) to determine the nearest neighbors. It can be useful when the data has different weights or when you need to take into account the nearest neighbors that are closer in the feature space.
2. LightGBMXT and LightGBM:
    2.1. LightGBM: LightGBM is a gradient boosting algorithm that is based on decision trees. It is characterized by a high learning speed and efficiency due to the use of a histogram approach to select divisions in trees.

- 2.2. LightGBMXT: LightGBMXT is an extension of LightGBM with support for distributed computing. It allows you to process large data sets and perform training on multiple computers, accelerating the training and prediction process.
- 2.3. LightGBMLarge: LightGBMLarge is an enhanced version of LightGBM specifically optimized for handling large data sets. It can process data that does not fit in RAM and supports distributed computing on multiple computers.
3. CatBoost:
    - 3.1. CatBoost is a gradient boosting algorithm that has been specifically designed to work with categorical features. It automatically processes categorical data without the need for pre-coding, which simplifies the data preparation process. CatBoost creates a prediction model in the form of an ensemble of weak prediction models, usually decision trees, has a mechanism for automatic processing of missing values and support for parallel processing.
4. XGBoost:
    - 4.1. XGBoost: XGBoost is a gradient boosting algorithm that is also based on decision trees. It differs from LightGBM in that it takes a different approach to separation selection and loss function optimization. XGBoost has many parameters for setting up the model and is widely used in data analysis and machine learning competitions.
5. RandomForestGini and RandomForestEntr:
    - 5.1. RandomForestGini: Random Forest is an ensemble model consisting of several decision trees. RandomForestGini uses the Gini index to select the division when building trees. It measures uncertainty in the data and seeks to reduce it with each separation.
    - 5.2. RandomForestEntr: This is also a random forest model, but instead of the Gini index, entropy is used to select the separation. Entropy measures the degree of uncertainty in the data, and minimizing entropy results in cleaner and more informative trees.
6. ExtraTreesGini and ExtraTreesEntr:
    - 6.1. ExtraTreesGini: ExtraTrees (Extremely Randomized Trees) is a random forest option that additionally randomly selects divisions when building trees. ExtraTreesGini uses the Gini index to select the partition and creates several random partitions to reduce the impact of outliers and noise in the data.
    - 6.2. ExtraTreesEntr: This is also an ExtraTrees model, but uses entropy to select separation in trees. Like ExtraTreesGini, it builds several random partitions to reduce the effect of noise and emissions.
7. NeuralNetFastAI and NeuralNetTorch:
    - 7.1. NeuralNetFastAI: NeuralNetFastAI is a neural network model based on the FastAI library. FastAI provides high-level tools for training deep neural networks, making the process easier and more intuitive.
    - 7.2. NeuralNetTorch: NeuralNetTorch uses the PyTorch framework to build and train neural networks. PyTorch provides flexibility and powerful tools for researching and developing deep learning models.
8. WeightedEnsemble_L2:
    - 8.1. WeightedEnsemble_L2: WeightedEnsemble_L2 is an ensemble model that combines the predictions of several basic models using weights. It uses the second level (L2) of the ensemble to combine the predictions of the underlying models and obtain the final forecast. Weights can be adjusted to optimize the quality of the ensemble.

ML models of the first level will be included in the model of the second level WeightedEnsemble_L2 if their accuracy on the training sample is equal to or greater than $k$, where $k$ is determined by the AutoML algorithm. AutoML model training will be performed using k-fold cross-validation on n times on n different random partitions of input data, and these parameters are selected automatically.

## Experimental results and discussion

The following technologies were used to implement the proposed framework in software and perform computational experiments.

Jupyter Notebook was used to create and execute code, as well as to document the process of analyzing data and building a model. Google Colaboratory, an online service that provides a free Jupyter Notebook runtime environment in the cloud, with the ability to use compute-intensive graphics processing units (GPUs) or tensor processing units (TPUs). For the development and implementation of all stages of the software project, the Python programming language was used along with the libraries:

   (1) **numpy.** A library for processing data arrays.
   (2) **pandas.** A library for working with data, including loading, preprocessing, and analysis.
   (3) **seaborn.** A library for visualizing data and creating informative graphs.
   (4) **matplotlib.pyplot**. A library for creating static, animated, and interactive data visualizations in Python.
   (5) **scikit-learn.** A machine learning library that provides a variety of algorithms, models, and metrics for building and evaluating models.
   (6) **autogluon.** A library that offers a set of libraries for performing automated machine learning, including automatic model fitting and hyperparameters.
   (7) **autogluon.tabular**. A library that provides automatic machine learning and model selection for data analysis tasks presented in the form of tables.

Let's consider the results and classification accuracy obtained by WeightedEnsemble_L2 model on the CVD dataset in three data preprocessing scenarios.

In the first scenario, S1_WeightedEnsemble_L2 model included two models, NeuralNetTorch and LightGBMLarge. Its accuracy on the training sample was 87.8%, and on the test sample - 88.81% (see Table 1).

**Table 1.** Comparison of AutoML models S1_WeightedEnsemble_L2 and ML models when using preprocessing of the data set according to scenario 1.

| № | model | Stack level | metrics | | | | | |
|---|---|---|---|---|---|---|---|---|
| | | | accuracy (test) | accuracy (val) | roc_auc | precision | f1 | recall |
| 1 | LightGBMXT | 1 | 0,8881 | 0,8522 | 0,9377 | 0,8750 | 0,8974 | 0,9211 |
| 2 | CatBoost | 1 | 0,8811 | 0,8348 | 0,9405 | 0,8642 | 0,8917 | 0,9211 |
| 3 | NeuralNetTorch | 1 | 0,8811 | 0,8522 | 0,9305 | 0,8831 | 0,8889 | 0,8947 |
| 4 | XGBoost | 1 | 0,8811 | 0,8348 | 0,9379 | 0,8734 | 0,8903 | 0,9079 |
| 5 | ExtraTreesEntr | 1 | 0,8811 | 0,8261 | 0,9302 | 0,8642 | 0,8917 | 0,9211 |
| 6 | LightGBM | 1 | 0,8741 | 0,8522 | 0,9313 | 0,8718 | 0,8831 | 0,8947 |
| 7 | ExtraTreesGini | 1 | 0,8741 | 0,8435 | 0,9358 | 0,8625 | 0,8846 | 0,9079 |
| 8 | LightGBMLarge | 1 | 0,8671 | 0,8609 | 0,9195 | 0,8519 | 0,8790 | 0,9079 |
| 9 | RandomForestGini | 1 | 0,8671 | 0,8174 | 0,9272 | 0,8608 | 0,8774 | 0,8947 |
| 10 | RandomForestEntr | 1 | 0,8601 | 0,8174 | 0,9353 | 0,8500 | 0,8718 | 0,8947 |
| 11 | NeuralNetFastAI | 1 | 0,8252 | 0,8348 | 0,8879 | 0,8148 | 0,8408 | 0,8684 |
| 12 | KNeighborsDist | 1 | 0,6853 | 0,7043 | 0,7251 | 0,7123 | 0,6980 | 0,6842 |
| 13 | KNeighborsUnif | 1 | 0,6573 | 0,6957 | 0,7135 | 0,6901 | 0,6667 | 0,6447 |
| 14 | S2_WeightedEnsemble_L2 | 2 | 0,8881 | 0,8783 | 0,9307 | 0,8846 | 0,8961 | 0,9079 |

In the second data preprocessing scenario, S2_WeightedEnsemble_L2 model included only one CatBoost model (see Fig.1), and the accuracy of the AutoML model improved compared to S1_WeightedEnsemble_L2: accuracy = 91.3% on the training sample, and accuracy = 92.3% on the test sample.

```
'stacker_info': {'num_base_models': 1, 'base_model_names': ['CatBoost']},

'model_weights': {'CatBoost': 1.0}
```

**Fig. 1.** The results of choosing an ML model in an ensemble S2_WeightedEnsemble_L2

Below is Table 2, which shows the main metrics of the model S2_WeightedEnsemble_L2 and ML models that it considered for inclusion in its weighted ensemble. As you can see, the model based on boosting decision trees turned out to be the best on the training sample, while the model was the best on the test sampleS2_WeightedEnsemble_L2 loses exactly to the models of the decision tree class, since they are focused on processing the categorical data that encoded all the values of the original data set in Scenario 2.

**Table 2.** Comparison of AutoML models S2_WeightedEnsemble_L2 and ML models when using preprocessing of the data set according to scenario 2.

| № | model | stack_level | Metrics | | | | | |
|---|---|---|---|---|---|---|---|---|
| | | | accuracy (test) | accuracy (val) | roc_auc | precision | f1 | recall |
| 1 | ExtraTreesEntr | 1 | 0,9650 | 0,8957 | 0,9863 | 0,9706 | 0,9635 | 0,9565 |
| 2 | ExtraTreesGini | 1 | 0,9580 | 0,8783 | 0,9863 | 0,9565 | 0,9565 | 0,9565 |
| 3 | RandomForestEntr | 1 | 0,9510 | 0,8783 | 0,9849 | 0,9559 | 0,9489 | 0,9420 |
| 4 | RandomForestGini | 1 | 0,9510 | 0,8870 | 0,9867 | 0,9559 | 0,9489 | 0,9420 |
| 5 | CatBoost | 1 | 0,9231 | 0,9130 | 0,9634 | 0,9265 | 0,9197 | 0,9130 |
| 6 | LightGBMXT | 1 | 0,8951 | 0,9043 | 0,9548 | 0,9219 | 0,8872 | 0,8551 |
| 7 | NeuralNetTorch | 1 | 0,8951 | 0,8957 | 0,9477 | 0,9219 | 0,8872 | 0,8551 |
| 8 | LightGBM | 1 | 0,8951 | 0,9043 | 0,9548 | 0,9219 | 0,8872 | 0,8551 |
| 9 | XGBoost | 1 | 0,8881 | 0,8783 | 0,9238 | 0,9077 | 0,8806 | 0,8551 |
| 10 | LightGBMLarge | 1 | 0,8601 | 0,8609 | 0,9555 | 0,9298 | 0,8413 | 0,7681 |
| 11 | NeuralNetFastAI | 1 | 0,8322 | 0,8609 | 0,9248 | 0,7922 | 0,8356 | 0,8841 |
| 12 | S2_WeightedEnsemble_L2 | 2 | 0,9231 | 0,9130 | 0,9634 | 0,9265 | 0,9197 | 0,9130 |

Interestingly, in the second scenario, two ML models were excluded by the WeightedEnsemble_L2 algorithm (see FIG.2), presumably due to the fact that they could not improve the final results of the weighted WeightedEnsemble_L2 model compared to the CatBoost model.

```
Fitting 13 L1 models ...
Fitting model: KNeighborsUnif ...
        No valid features to train KNeighborsUnif... Skipping this model.
Fitting model: KNeighborsDist ...
        No valid features to train KNeighborsDist... Skipping this model.
```

**Fig.2.** Automatic skipping of two machine learning models

The accuracy of predicting CVDs using WeightedEnsemble_L2 model in which the third data preprocessing scenario was used is presented in Table 3.

**Table 3**. Comparison of AutoML models S3_WeightedEnsemble_L2 and ML models when using preprocessing of the data set according to scenario 3.

| № | model | stack_level | metrics | | | | | |
|---|---|---|---|---|---|---|---|---|
| | | | accuracy (test) | accuracy (val) | roc_auc | precision | f1 | recall |
| 1 | NeuralNetFastAI | 1 | 0,8811 | 0,9478 | 0,9151 | 0,8308 | 0,8640 | 0,9000 |
| 2 | LightGBMXT | 1 | 0,8741 | 0,9478 | 0,9267 | 0,8281 | 0,8548 | 0,8833 |
| 3 | ExtraTreesEntr | 1 | 0,8741 | 0,9217 | 0,9241 | 0,8281 | 0,8548 | 0,8833 |
| 4 | ExtraTreesGini | 1 | 0,8671 | 0,9304 | 0,9271 | 0,8154 | 0,8480 | 0,8833 |
| 5 | XGBoost | 1 | 0,8531 | 0,9478 | 0,9086 | 0,8000 | 0,8320 | 0,8667 |
| 6 | RandomForestEntr | 1 | 0,8531 | 0,9304 | 0,9173 | 0,8000 | 0,8320 | 0,8667 |
| 7 | RandomForestGini | 1 | 0,8531 | 0,9217 | 0,9143 | 0,8095 | 0,8293 | 0,8500 |
| 8 | CatBoost | 1 | 0,8322 | 0,9391 | 0,9303 | 0,7903 | 0,8033 | 0,8167 |
| 9 | NeuralNetTorch | 1 | 0,8322 | 0,9304 | 0,9315 | 0,7903 | 0,8033 | 0,8167 |
| 10 | LightGBMLarge | 1 | 0,8252 | 0,9304 | 0,9072 | 0,7692 | 0,8000 | 0,8333 |
| 11 | LightGBM | 1 | 0,8182 | 0,9304 | 0,9096 | 0,7576 | 0,7937 | 0,8333 |
| 12 | KNeighborsUnif | 1 | 0,7133 | 0,7565 | 0,7599 | 0,6610 | 0,6555 | 0,6500 |
| 13 | KNeighborsDist | 1 | 0,7063 | 0,7565 | 0,7646 | 0,6552 | 0,6441 | 0,6333 |
| 14 | S3_WeightedEnsemble_L2 | 2 | 0,8741 | 0,9565 | 0,9259 | 0,8281 | 0,8548 | 0,8833 |

As follows from Table 3, at the training stage S3_WeightedEnsemble_L2 the model showed an accuracy of 95.65%, and on the test set, its accuracy was 87.41%. Two of the 13 models, namely LightGBMXT and ExtraTreesGini, were automatically included in its weighted ensemble.

The conducted studies of the effectiveness of the detection of CVDs using models allows us to formulate answers to research questions:

RQ1. Will the use of an AutoML model that includes optimized baseline ML models lead to an increase in the accuracy of detecting CVDs compared to basic ML models on a test sample? In the first and third scenarios, the accuracy of the AutoML model is comparable to the best optimized ML models. In the second scenario, the AutoML model does not exceed the accuracy of several basic optimized models.

RQ2. Can AutoML's built-in preprocessing algorithm compete in efficiency with data preprocessing algorithms created by a data scientist? The built-in AutoML preprocessing algorithm was used in the first scenario, the accuracy on the test sample demonstrated by the AutoML model was lower than in the second scenario and slightly higher than in the third data preprocessing scenario.

**Conclusion**

Automated machine learning makes it possible to simplify and speed up the process of developing AI/ML applications, which is key in the development of POS by application users, in particular medical specialists. This article explored the capabilities of AutoML using the example of one WeightedEnsemble_L2 model and 13 basic ML models for predicting cardiovascular diseases for 11 patient health indicators, using a database combining observations of five open data sets from the UCI Machine Learning Repository (https://archive.ics.uci.edu/ml/machine-learning-databases/heart-disease/).

The authors' contribution lies in a new framework for exploring the capabilities and effectiveness of the AutoML model and ML models on a new database for detecting cardiovascular diseases. The results showed that, depending on the type of data preprocessing, in particular, on the technique of data normalization, the AutoML algorithm creates models with different structures that implement predictions of cardiovascular diseases with different accuracy. In our study, we used three data preprocessing scenarios: (S1) built-in AutoML data preprocessing, (S2) software-implemented data preprocessing with the transformation of values into categorical binary features, and (S3) software-implemented data preprocessing with z-transformation of continuous values into a range [0,1]. It was shown that the accuracy of the WeightedEnsemble_L2 model, which used the built-in data preprocessing in AutoML, was 87.8% on the training sample, and 88.81% on the test sample. While the generated model WeightedEnsemble_L2 based on the data prepared by the second scenario, predicted the presence of CVD on the training sample with an accuracy of 91.3%, and on the test sample with an accuracy of 92.3%. The AutoML model, the data of which was preprocessed according to the third scenario, demonstrated an accuracy of 95.65% on the training set, and its accuracy was 87.41% on the test set. Of course, the use of AutoML algorithms has a great future and diverse applications in healthcare. However, questions remain which AutoML model to choose when solving a specific clinic problem and how to explain the solution it obtains.

The authors declare that they have no known competing financial interests or personal relationships that could have appeared to influence the work reported in this paper.

**ACKNOWLEDGMENTS.** This research was performed in the framework of the state task in the field of scientific activity of the Ministry of Science and Higher Education of the Russian Federation, project "Models, methods, and algorithms of artificial intelligence in the problems of economics for the analysis and style transfer of multidimensional datasets, time series forecasting, and recommendation systems design", grant no. FSSW-2023-0004.